\documentclass[conference,letterpaper]{IEEEtran}
\IEEEoverridecommandlockouts
\usepackage{cite}      

\usepackage{booktabs}  
\usepackage{threeparttable}

\usepackage{graphicx}
\usepackage{subfig}

\usepackage{amsmath}   
\usepackage{multirow}
\usepackage[left=0.71in,top=0.94in,right=0.71in,bottom=1.18in]{geometry}
\begin{document}
	
	\title{\LARGE \bf
		A Robust Stereo Camera Localization Method with Prior LiDAR Map Constrains
	}
	\author{Dong Han$^{1,2}$, Zuhao Zou$^{1}$, Lujia Wang$^{1}$ and Cheng-Zhong Xu$^{3}$
	\thanks{$^{1}$ Dong Han, Zuhao Zou and Lujia Wang are with the Shenzhen Institutes of Advanced Technology, Chinese Academy of Sciences, Shenzhen, China. \{dong.han, zh.zou, lj.wang1\}{\tt\small @siat.ac.cn}}%
	\thanks{$^{2}$  Dong Han is also with the University of Chinese Academy of Sciences.}
	\thanks{$^{3}$  Cheng-Zhong Xu is with the University of Macau. czxu@um.edu.mo}
	\thanks{$^{*}$  This research is supported by the Shenzhen Science and Technology Innovation Commission (Grant Number JCYJ2017081853518789), the Guangdong Science and Technology Plan Guangdong-Hong Kong Cooperation Innovation Platform (Grant Number 2018B050502009) and the National Natural Science Foundation of China (Grant Number 61603376) awarded to Dr. Lujia Wang.}
	%
}


	\maketitle
\begin{abstract}
	In complex environments, low-cost and robust  localization is a challenging problem.
	For example, in a GPS-denied environment, LiDAR can provide accurate position information, but the cost is high. 
	In general, visual SLAM based localization methods become unreliable when the sunlight changes greatly. 
	Therefore, inexpensive and reliable methods are required. 
	In this paper, we propose a stereo visual localization method based on the prior LiDAR map. 
	Different from the conventional visual localization system, we design a novel visual optimization model by matching planar information between the LiDAR map and visual image.
	Bundle adjustment is built by using coplanarity constraints. To solve the optimization problem, we use a graph-based optimization algorithm and a local window optimization method.
	Finally, we estimate a full six degrees of freedom (DOF) pose without scale drift.
	To validate the efficiency, the proposed method has been tested on the KITTI dataset. The results show that our method is more robust and accurate than the state-of-art ORB-SLAM2.
	\\
\end{abstract}


\section{Introduction}
High-precision localization is necessary for autonomous vehicles. 
In past research, many kinds of sensors have been adopted for localization in a complex environment. 
The LiDAR is often considered the most reliable sensor in mapping and localization due to its accurate range measurements. 
However, its high cost is an obstacle for applications.
On the other hand, GPS can perform well in the intense signal area, but it may fail to provide accurate localization when in urban areas and indoor environment. 
Cameras have been proposed as a substitute for LiDARs because of its low cost, small size, and ability to get color information. 
However, a monocular camera suffers from a fatal weakness, scale uncertainty , which can cause angular drift. Stereo camera systems overcome the problem of scale uncertainty. However, their accuracy and robustness still do not catch up with those of LiDARs.\par 
Different situations have different requirements of precision and system cost\cite{lm_rcar_2016}.
Mapping requires high-precision sensors, in this work, it is less sensitive to price because the device can be reused.
For autonomous vehicles localization, as the number of vehicles increases, the number of sensors also increases.
Considering accuracy and cost, high-precision and low-cost sensors are needed, LiDAR is not considered at this time because of its price.   
\begin{figure}[t]
	\centering
	\includegraphics[scale=0.31]{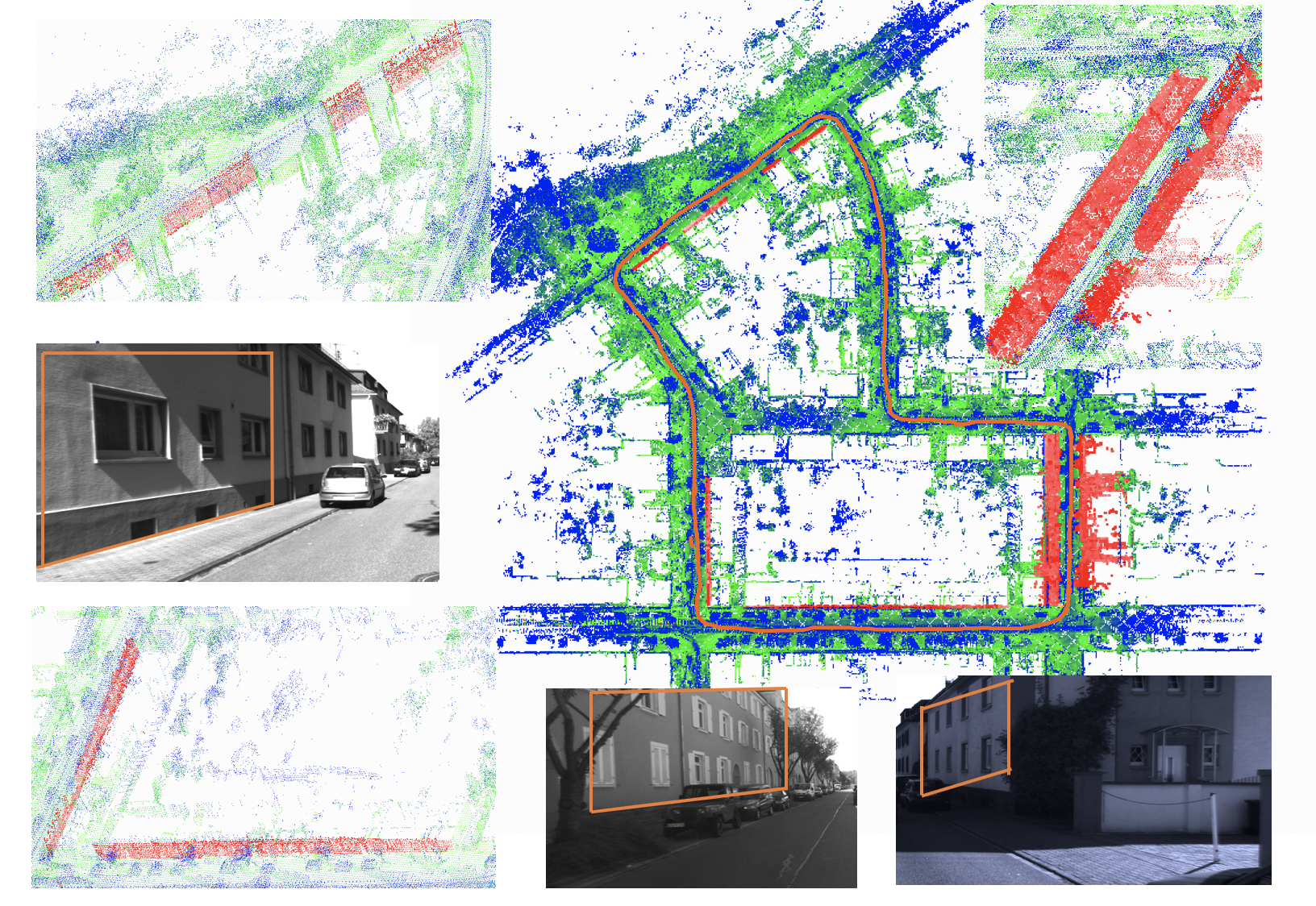}
	\caption{
		LiDAR map is produced by G-loam(GPS based Loam) using a Velodyne HDL-64E LiDAR. As shown on the LiDAR map, there are at least six planes; the green line marks plane information in the visual image and the red point is the plane in the LiDAR map. The the red trajectory shows the camera position in the point cloud map.
	}
	\label{point cloud map}
\end{figure}
An effective method to decrease cost and maintain precision is to combine the advantages of cameras and LiDARs. Generally, there are two ways to fuse LiDARs and cameras\cite{liu12dpfusion}. 
The first is to synthesize images from the LiDAR map\cite{steder2011place}, \cite{pascoe2015direct}. This method requires solving the relative extrinsic parameters of the camera and LiDAR, but the computation is heavy for registering images to the point cloud. 
Therefore, it is not suitable for a localization system, which has a strong requirement of real-time performance. 
The other way is to make landmarks both from the LiDAR map and visual image\mbox{\cite{wang20133d}, \cite{ding2018laser}}. In particular lane lines are the most common landmark to aid visual localization. 
However, this method only satisfies some \mbox{special} scenarios. 
Moreover, the lane line extraction is also a difficult problem. 
So, the above mentioned methods can not work well in the complex environment. 
These methods also need the LiDAR and camera to be calibrated, and the calibration result can affect the accuracy of the localization result. \par
To avoid sensor calibration, we propose a robust localization method, which only needs a stereo camera. Unlike fusing the point cloud and image directly, we extract geometric features from a prior LiDAR map generated by some algorithms like G-LOAM\cite{zheng2019low}, then pick out points with the same geometric properties in the visual image. These points satisfy bundle adjustment(BA) constraints as well as satisfy geometric constraints. Overall, the contributions of this paper are as follows:\par

\begin{itemize}
	\item We propose an accurate and robust stereo visual localization method, which only relies on a camera and a prior LiDAR map.
	\item We design a new visual optimization model based on bundle adjustment. 
	\item We propose a new framework for fusing camera and LiDAR, which greatly reduces the dependence on the LiDAR.
	
\end{itemize}\par
The rest of the paper is organized as follows: We discuss related work in \mbox{Section \uppercase\expandafter{\romannumeral2}}
, describe our method in Section \uppercase\expandafter{\romannumeral3}, and present the experimental results in Section \uppercase\expandafter{\romannumeral4}. Conclusions are given in Section \uppercase\expandafter{\romannumeral5}.

\section{Related Works}
Camera-\cite{5509441},\cite{liu12scale} and LiDAR-\cite{BhuttaPCR} based methods are common for mobile robot localization. Vision sensors capture the appearance of the surrounding environment, while LiDARs provide accurate range information and are mostly invariant to lighting conditions. In the past decades, great progress has been made in both vision-based localization and LiDAR-based localization. \par
For visual localization, there are two main approaches, feature-based methods\cite{davison2007monoslam},\cite{klein2007parallel} and direct methods\cite{newcombe2011dtam},\cite{engel2014lsd},\cite{forster2014svo}. Among feature-based methods, ORB-SLAM2\cite{mur2017orb} is a classical framework. In this method, ORB features are extracted from the image and BA is used, which minimizes the reprojection error over multiple image frames, to solve optimization problems. Direct methods, on the other hand, use an optical flow model to track motion points. In \cite{engel2017direct}, the authors proposed to minimize the photometric error by using a sliding window. For LiDAR localization, the most common approach is to rely on the intensity information of the surrounding environment, and with the help of lanes to obtain the position information\cite{hata2014road}. Meanwhile, \cite{wolcott2017robust} proposes a generic probabilistic method for localization. This algorithm uses Gaussians to model the world, which stores the z-height and intensity distribution of the environment. \par
Because the traditional vision-based  methods fails to meet the accuracy requirements of localization, while LiDAR-based method have a high cost map-based visual localization has been an active field of research in recent years.
\cite{wolcott2014visual} proposes to use a single monocular camera within a 3D prior ground-map. The map is generated by LiDAR and the height information is removed. The novelty of this work is using a GPU to generate several synthetic views from different poses, then calculating the normalized mutual information between the real camera measurements and these synthetic views, and finally finding the maximized NMI of the synthetic view. This view is the pose we need.
\cite{lu2017monocular} presents a monocular vision-based approach for localization in urban environments using road markings. First, a random forest-based edge detector is employed to detect road lanes. Then the Chamfer distance is computed between the detected edges and the projected road marking points in the image space. Epipolar geometry constraints and odometry are taken into account to formulate a non-linear optimization problem to estimate the six-DoF camera pose.
In \cite{kim2018stereo}, SGBM is used for estimating disparity, and recover the depth from stereo images. Depth from the stereo camera is matched with the prior LiDAR map. A full six degree of freedom camera pose is estimated via minimizing the depth residual. This method is based on ORB-SLAM2 which adds depth constraint when tracking. We are also interested in cloud robotic systems\cite{Lujia2012Towards},\cite{7403967},\cite{8772088},\cite{7060735} and we will apply our work to cloud robots in the future.
\section{METHOD}

\begin{figure}[t]
	\centering
	\includegraphics[scale=0.32]{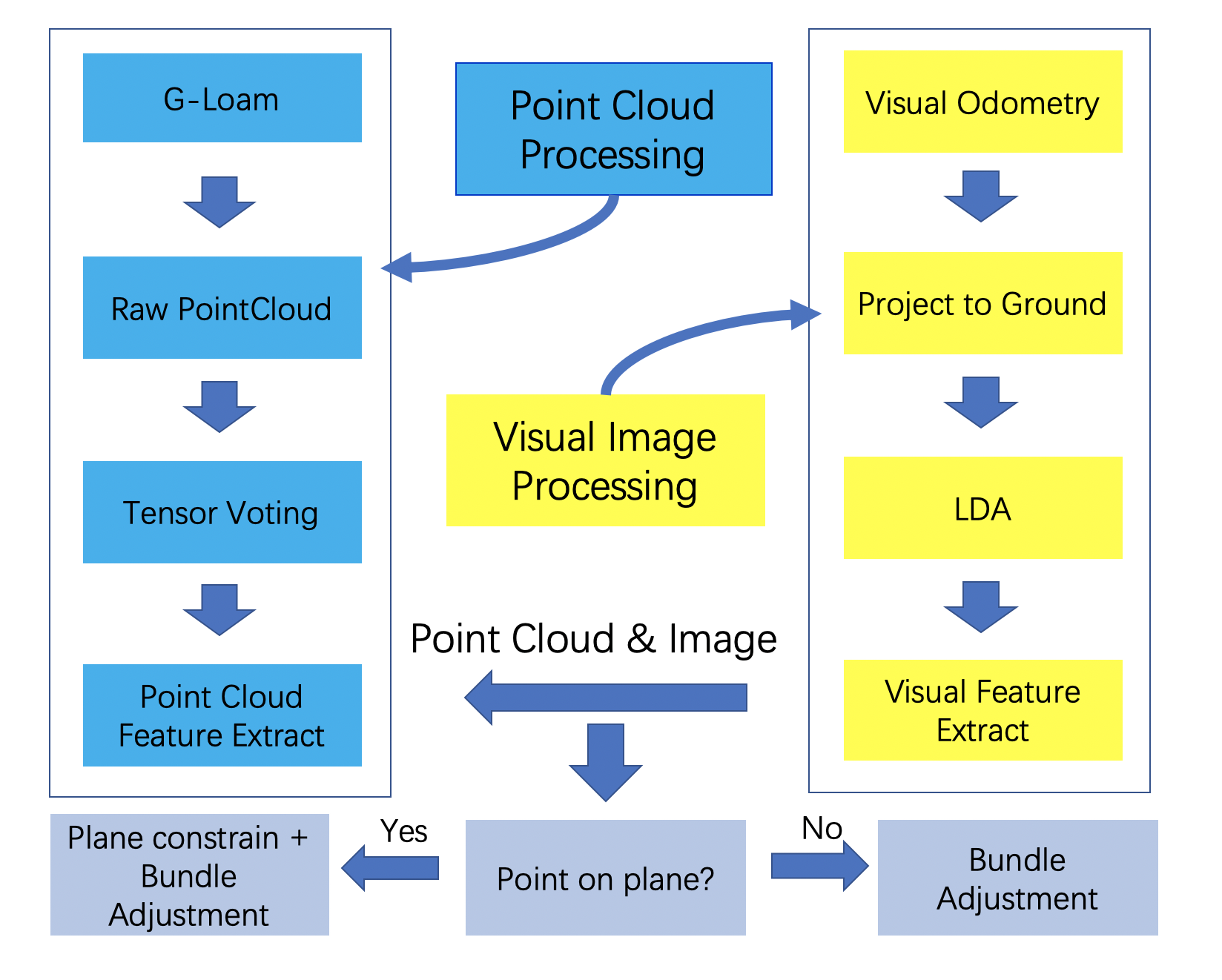}
	\caption{
		The system structure,  which includes three parts. The blue box shows the framework of point cloud processing, the yellow box shows the framework of visual image processing, and the purple box shows the framework of visual localization.  
	}
\end{figure}
\par
\subsection{Problem Definition}
In this paper, the camera frame is denoted as $\mathbf{F}_{camera}$ and the prior map from the LiDAR as $\mathbf{M}_{LiDAR}$. We represent the pose of the camera  as  $\mathbf{T} \in \mathbf{SE(3)}$, which transforms a point $\mathbf p \in R^3$ in the current frame to world coordinate  . For the sake of convenience in the computation, we map $\mathbf{T}$ to $\mathbf{\xi} \in \mathbf {se(3)}$ by using operators $\mathbf{Log(.)}$. $\mathbf{R} \in \mathbf {SO(3)} $ and $\mathbf{t} \in\mathbf R^3$ represent the rotation matrix and translation vector respectively. We use a stereo camera model, and the intrinsic of the camera $ \mathbf K $ is given. The problem is: input $\mathbf{M}_{LiDAR}$ and $\mathbf T_{camera}$ and output a more accurate $\mathbf T_{current}$.

\subsection{System Overview}

The framework of our method contains three parts, a stereo online visual localization, and an offline point cloud map processing method. The high-accuracy prior map is generated by using GPS and LiDAR. In the framework of offline point cloud map processing, we use Tensor Voting to extract plane features form the map $\mathbf{M}_{LiDAR}$ and use K-means to get the normal of the surface. \par
As stereo images input, we can get the depth information by using triangulation measurement. So, our system first initializes visual localization with a frame of the picture. After initialization, the ORB-SLAM2 tracking thread is employed in visual localization. During back- end optimization, ORB-SLAM2 mainly uses bundle adjustment to minimize the reprojection error. Now, since the plane information has been obtained, the point on plane satisfied BA constrain, but plane constrains. As shown in Fig. 3, the all point fall into two categories, one class is not on the plane, because we use pinhole camera model, these points satisfy pinhole camera constraint, the other class point satisfy pinhole camera constraint and plane constrain.\par
\subsection{Point Cloud Map Processing}
Using LiDAR and GPS to generate high-precision maps, it is necessary to extract useful geometric information from the map to provide constraints for visual positioning. From observation, we find that the plane feature is a piece of beneficial information in the urban environment, and the plane is relatively easy to extract in point cloud maps. \par
The point cloud map processing is divided into two steps. The first step is to use the Tensor Voting\cite{medioni2000tensor},\cite{liu2014efficient},\cite{liu12robio} algorithm to solve the normal vector of each point in the point cloud. In the second step, the k-means algorithm is used to cluster points with the same normal, and the plane equation is solved according to the results of clustering.\par
Tensor Voting is an effective method for solving the surface normal. The essential idea is to extract implicit geometric features from a large amount of scattered point cloud data by transferring tensors between adjacent points. With the increase of distance, the influence coefficient of points in voting field decays gradually. According to this principle, we set the tensor kernel as
\begin{equation}\label{equ1}
Decay(d,\sigma)=e^{-\frac{d}{{\sigma}^2}},
\end{equation}
where $d={(x_i-x)}^2+{(y_i-y)}^2$, $(x,y)$ represents the coordinates of the votee point, and $(x_i,y_i)$ represents the coordinates of the voter point, and $\sigma$ is the kernel size of the sparse voting field. 
The input point $\mathbf P$ can be expressed by the second-order symmetric semi-positive definite tensor $\mathbf T$. Because of the equivalent relation between tensor and matrix, $\mathbf T_{3\times 3}$ can be expressed by a positive semidefinite matrix. and be decomposed into three parts.\par
$\mathbf T_{3\times 3}$ can be decomposed into the following forms:
\begin{align}
T&=\lambda_1e_1e_1^T+\lambda_2e_2e_2^T+\lambda_3e_3e_3^T\\
&=(\lambda_1-\lambda_2)e_1e_1^T\\
&+(\lambda_2-\lambda_3)(e_1e_1^T+e_2e_2^T)\\
&+\lambda_3(e_1e_1^T+e_2e_2^T+e_3e_3^T).
\end{align}

 In (2), $T$ is decomposed into three parts,(3)  describes a stick, (4)  describes a plate, and (5) describes a ball. $\lambda_1 \geq \lambda_2 \geq \lambda_3 \geq0$ is the eigenvalue of $\mathbf T$, $\mathbf {e_1},\mathbf {e_2},\mathbf {e_3}$ is corresponding eigenvectors. \par
According to the saliency of each tensor component, scattered point clouds are divided into three categories:\par
\begin{itemize}
	\item[    (1)] if $(\lambda_1-\lambda_2)\geq(\lambda_2-\lambda_3)$ and $(\lambda_1-\lambda_2)\geq \lambda_3$, the point belongs to the stick, and the orientation is $e_1$;
	\item[    (2)] if $(\lambda_2-\lambda_3)\geq(\lambda_1-\lambda_2)$ and $(\lambda_2-\lambda_3)\geq \lambda_3$,
	the point belongs to the plate, and the orientation is $e_3$.
	\item[    (3)] if $(\lambda_3>>(\lambda_1-\lambda_2)$ and $\lambda_3>>(\lambda_2-\lambda_3)$, the point belongs to the ball, and the orientation is uncertain.
\end{itemize}\par
After the tensor voting framework, we can get the normal of each point. From (2), we can find the points on the wall conform to the following characteristics:
\begin{equation}
\mathbf {T_p}=(\lambda_2-\lambda_3)(\mathbf {e_1e_1^T}+\mathbf {e_2e_2^T}).
\end{equation}\par
K-means algorithm is employed for clustering point clouds. The surface normal can be easily obtained, and then the plane equation can be acquired.


\subsection{Visual Image Processing}

For point cloud data generated by the camera, due to the sparsity of the point cloud, we are searching for line information in two-dimensional space. We have experimented with a variety of methods for line detection. The first method uses the Hough transform for line detection. However, the Hough transform depends on parameter adjustment and is not suitable for the complex and changeable environments. Another method is to use LSD \cite{von2008lsd} for line detection. The algorithm does not depend on parameter changes, and the detection speed is faster than the Hough transform\cite{ballard1981generalizing}.\par

As we know, linear detection mainly depends on detecting the pixels with large gradient changes, so LSD is mainly used to detect the local straight contours in images, in which there are sharp changes from black to white or from white to black. Firstly, the image gradient is calculated, then the gradient of each pixel is sorted, and the gradient of points is used for local region growth. Finally, the similar gradient points are clustered, and the straight-line part of the image map is obtained.\par

\subsection{Visual Localization}

Traditionally, visual localization is divided into three stages. Firstly, the ORB feature between two frames is matched to calculate the pose between two frames. Then, the depth information is restored by triangulation measurement. Finally, Bundle Adjustment is done according to the map point and corresponding frames\cite{zhang2019vr},\cite{liu12navigation}.\par
In this paper, our method is to add projection constraints for the points on the wall in addition to bundle adjustment when optimizing. In the initialization stage of localization, the original BA constraints are only relied on because there is too little valid information to be relied on. With the increase of input valid information, the number of optimization variables increases gradually, so the output pose is more accurate.\par
In the visual image processing framework, we can get the line feature of the local map. For the current observation points, the observed points are used to provide a prior constraint for the current point. This constraint function can be expressed by:
\begin{equation}
E=\sum_{k=1}^{K} \sum_{j\in F(k)} \lambda E_{ba}+ \sum_{p\in \pi_i} (1-\lambda)E_{projection}
\end{equation}
\begin{figure}[t]
	\centering
	\includegraphics[scale=0.41]{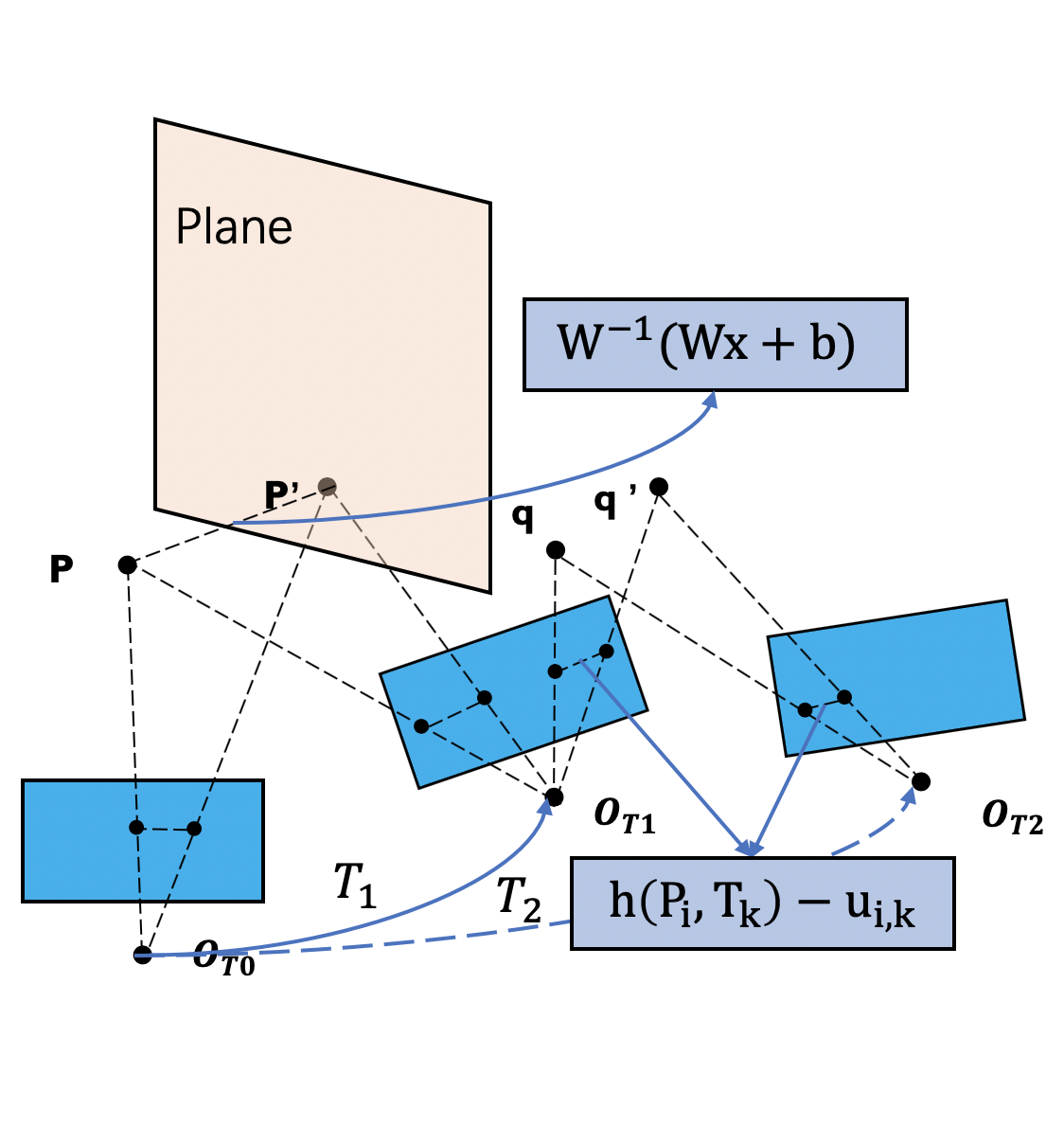}
	\caption{
		Visual points can be divided into two categories, point $\mathbf{p}$ is on the plane and point $\mathbf{q}$ is not on the plane. In the visual localization framework, point $\mathbf{p}$ needs to minimize the reprojection error and coplanarity error and point $\mathbf{q}$ is required to minimize the reprojection error.
	}
\end{figure}
Here, we use the incremental sliding window method to connect the observed points with the current point. In each incoming frame, the target points in this frame are optimized simultaneously with the observed images to provide a maximum six-DOF pose. The most easily lost location information is at the corner, which corresponds to the end of each street. Therefore, at each corner, the sliding window will reach the maximum value, which provides more information for the constraints of the next frame, and also eliminate the problem of location drift.

\subsubsection{No Additional Constraints}
In optimization problems without additional constraints, bundle adjustment is mainly used to solve optimization problems. We take the Lie algebra corresponding to $\mathbf{R}$ and $\mathbf{t}$ as $\mathbf{\xi}$. $\mathbf p$ represents the observed map point. 
The above formula represents the error caused by the observation of the $k^{th}$ point in the $j^{th}$ frame. The cost function is as follows:
\begin{equation}
E_{ba}=||(u_i-Kexp(\xi\hat \ )p_k \ )^TQ^{-1}_{k,j}(u_i-Kexp(\xi\hat \ )p_k \ )||_2^2.
\end{equation}
\subsubsection{Additional Constraints}
For the points on the wall, the sliding window algorithm is used to add plane constraints.
In the first part of the algorithm, only the BA algorithm is used to constrain it. With the increase of input plate information, the points on the wall gradually increase, and the constraints of the points on the wall are added. Because of the increase in the points, the precise pose can be obtained. The plane equation is $Wx+b=0$, so the projection cost function is shown as follows:
\begin{equation}
E_{projection}=||(\frac{1}{|W|}(Wp_j+b))^TR^{-1}_{j}(\frac{1}{|W|}(Wp_j+b))||_2^2.
\end{equation}

To solve the optimization problem in visual localization, we adopt a graph-based method, which is generally used in solving the SLAM problem. We establish a graph-model based on an optimization problems, and in a graph, each edge represents different constraints. BA constraint, and plane constraints are increased to our system.

\begin{figure}[t]
	\includegraphics[scale=0.35]{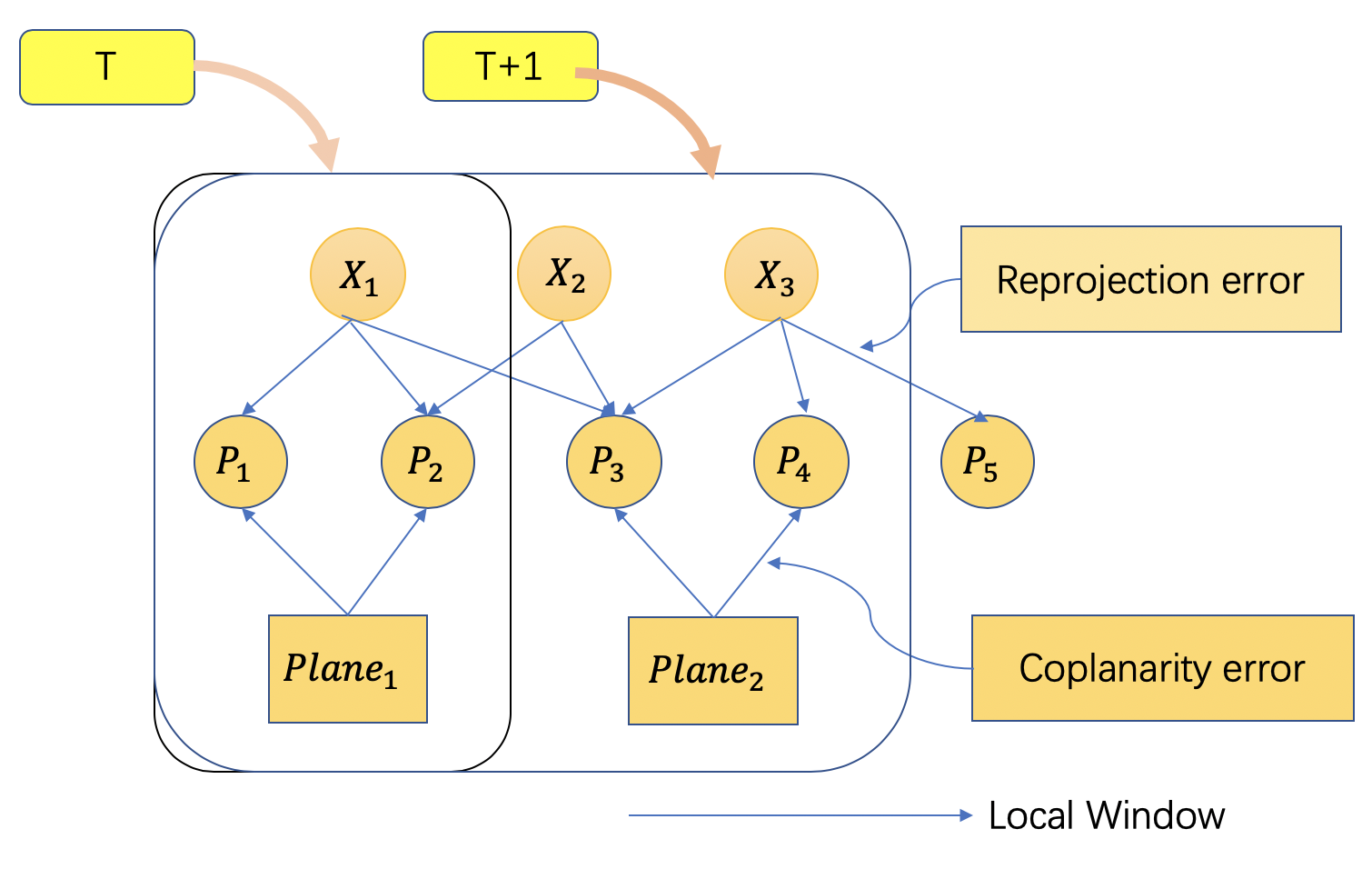}
	\caption{
		Graph-structure optimization. $\mathbf{P}$ is the node of visual point, and $\mathbf{X}$ represent the pose of frame, $\mathbf{Plane}$ is the plane information extracted from LiDAR map.
	}
\end{figure}
We let $\mathbf{P}$ be the node of visual point, and $\mathbf{X}$ represent the pose of frame, $\mathbf{P_L}$ is the point on plane which we extract form LiDAR map. We also define the error function $\mathbf{e_k(p_k,T)}$ between $\mathbf {p_k}$ and observation point, we use a edge to represent it. $\mathbf{e_l(p_k,p_l)}$ is defined as the cost function which project point to the plane, same as  $\mathbf{e_k(p_k,T)}$, we also use a edge to represent it. We can use (10) to describe  optimization problem: 
\begin{align}
F(p)=&\sum_{k=1}^{K} \sum_{j\in F(k)}e_l(p_k,p_l)^TQ^{-1}_{k,j}e_l(p_k,p_l)\nonumber \\
&+ \sum_{p\in \pi_i} e_l(p_k,p_l)^TR^{-1}_{j}e_l(p_k,p_l).
\end{align}
After get (10),first we can expand it by using taylor expansion,
\begin{equation}
F(p+\delta p)=F(p)+J(p)\delta p,
\end{equation}
$J(p)$ is Jacobian matrix of $F(p)$ which is a sparse matrix, We use Levenberg-Marquadt algorithm to solve the problem, our aim is to chose appropriate $p$ and $T$ to minimize error and make $F(p)=\alpha$. So the problem will turn into solve (10):
\begin{equation}
(J^TJ+\lambda I)\delta x=-J^T\alpha.
\end{equation}

To solve this graph optimization problem, Ceres is employed to solve the equation, which is an open-source C++ library for modeling and solving large, complicated optimization problems.
\section{Experiment}
We test our algorithm on the KITTI dataset and compare our framework with ORB-SLAM2. KITTI-07 is an outdoor image sequence that includes 1101 stereo images,
and this dataset is based on the urban scene. The experiment is divided into two parts, which include mapping and localization.\par
Our localization algorithm is based on the prior map generated by LiDAR and GPS. So the first step is to acquire a high-precision map. The KITTI odometry dataset provides a sequence of Velodyne HDL-64E LiDAR scans; we use this dataset and G-Loam mapping algorithm to produce a 3D LiDAR Map.  The GPS-INS system provides the ground truth of the camera pose. As shown in Fig. 1, the red line is the trajectory of ground true.\par
For localization, we test ORB-SLAM2 on the same dataset, and because we only test the localization model, the loop closing function in ORB-SLAM2 is closed. 
\begin{figure}[t]
	\includegraphics[scale=0.553]{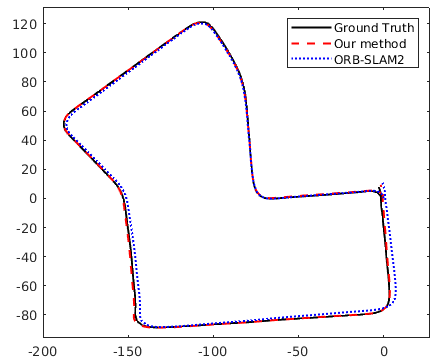}
	\caption{
		Camera trajectory on the KITTI-07 dataset. The blue line is the result of ORB-SLAM2, the red line is that of our method, and the black line is the ground truth. 
	}
\end{figure}
\par
As shown in Fig. 5. Our method is more accurate than ORB-SLAM2 localization, especially in the corner of the street, ORB-SLAM2 easily lost the position and is not able to correct the pose; it will cause error accumulation. And our method is more robust, even if in the corner because we use the plane constraint to aid visual localization, it eliminates error.\par
\renewcommand{\arraystretch}{1.5} 
\begin{table}[tp]  
	\centering  
	\fontsize{6.5}{8}\selectfont  
	\begin{threeparttable}  
		\caption{Translation error result from KITTI-07 }  
		\label{tab:performance_comparison}  
		\begin{tabular}{ccccccc}  
			\toprule  
			\multirow{2}{*}{Method}&  
			\multicolumn{3}{c}{ Our Method}&\multicolumn{3}{c}{ ORBSLAM2}\cr  
			\cmidrule(lr){2-4} \cmidrule(lr){5-7}  
			&Mean&RMSE&Std&Mean&RMSE&Std\cr  
			\midrule  
			KITTI 07&0.53264&0.5643&0.2031&0.7535&0.8433&0.3786\cr  
			\bottomrule  
		\end{tabular}  
	\end{threeparttable}  
\end{table}  
In order to evaluate our algorithm, the statistics of absolute trajectory error(ATE) are computed, and ATE figure is drawn. We compute RMSE(root-mean-square error), Mean, Std(Standard Deviation), and each indicator is better than ORB-SLAM2. As shown in the table, our method is more robust than ORB-SLAM2. We calculate the error by this follow form $e(t)=\sqrt{ e(t)_x^2+e(t)_y^2}$. From the result, we can get the average error is 0.5326m, and the maximum error is 0.9664m, minimum error is 0.1884. About ORB-SLAM2, the average error is 0.8060, and the maximum error is 2.0300m, minimum error is 0.1276m.

\section{Conclution And Future Work}

In this paper, we propose a stereo visual localization method based on the prior LiDAR map. 
We only use a stereo camera to acquire decimeter-level localization. Compare with localization based on LiDAR, we cost low and reach the same level of precision.
Our method performs robust in a complex environment, which can provide the accurate estimation of a six-DoF pose in urban without GPS signal. It can also work well in a sunlight change environment and without scale drift.\par

To combine the advantages of LiDAR and camera and cut costs, we design a novel visual optimization model by matching planar information between LiDAR map and visual image. To achieve the real-time and robust localization, LDA and Tensor Voting are employed to extract geometric features in visual image and point cloud map, respectively. We use the coplanarity constrains to build bundle adjustment and solve it using graph-based optimization algorithm a local window optimization method.\par
In the experiment part, we test our approach in the KITTI urban environment. The result shows our method is more robust and accurate than ORB-SLAM2. This result proves that our method has a great advantage in this environment.
In the future, we will try to extract more geometry features from the stereo image, increase line constrain to localization. Furthermore, we will implement evaluation in the long term localization application.

	\bibliography{db}{}

\begin{thebibliography}{10}

\bibitem{lm_rcar_2016}
I.~Deutsch, M.~Liu, and R.~Siegwart, ``A framework for multi-robot pose graph
  slam,'' in {\em Real-time Computing and Robotics (RCAR) 2016 IEEE
  International Conference on}, (Angkor Wat, Cambodia), June 2016.

\bibitem{liu12dpfusion}
M.~Liu, L.~Wang, and R.~Siegwart, ``{DP-Fusion: A generic framework for online
  multi sensor recognition},'' in {\em IEEE Conference on Multisensor Fusion
  and Integration for Intelligent Systems (MFI)}, IEEE, 2012.

\bibitem{steder2011place}
B.~Steder, M.~Ruhnke, S.~Grzonka, and W.~Burgard, ``Place recognition in 3d
  scans using a combination of bag of words and point feature based relative
  pose estimation,'' in {\em 2011 IEEE/RSJ International Conference on
  Intelligent Robots and Systems}, pp.~1249--1255, IEEE, 2011.

\bibitem{pascoe2015direct}
G.~Pascoe, W.~Maddern, and P.~Newman, ``Direct visual localisation and
  calibration for road vehicles in changing city environments,'' in {\em
  Proceedings of the IEEE International Conference on Computer Vision
  Workshops}, pp.~9--16, 2015.

\bibitem{wang20133d}
R.~Wang, ``3d building modeling using images and lidar: A review,'' {\em
  International Journal of Image and Data Fusion}, vol.~4, no.~4, pp.~273--292,
  2013.

\bibitem{ding2018laser}
X.~Ding, Y.~Wang, D.~Li, L.~Tang, H.~Yin, and R.~Xiong, ``Laser map aided
  visual inertial localization in changing environment,'' in {\em 2018 IEEE/RSJ
  International Conference on Intelligent Robots and Systems (IROS)},
  pp.~4794--4801, IEEE, 2018.

\bibitem{zheng2019low}
L.~Zheng, Y.~Zhu, B.~Xue, M.~Liu, and R.~Fan, ``Low-cost gps-aided lidar state
  estimation and map building,'' {\em arXiv preprint arXiv:1910.12731}, 2019.

\bibitem{5509441}
M.~Liu, C.~Pradalier, Q.~Chen, and R.~Siegwart, ``A bearing-only 2d/3d-homing
  method under a visual servoing framework,'' in {\em 2010 IEEE International
  Conference on Robotics and Automation}, pp.~4062--4067, May 2010.

\bibitem{liu12scale}
M.~Liu, C.~Pradalier, F.~Pomerleau, and R.~Siegwart, ``{Scale-only Visual
  Homing from an Omnidirectional Camera},'' in {\em Proc. of the IEEE
  International Conference on Robotics and Automation (ICRA)}, 2012.

\bibitem{BhuttaPCR}
M.~U.~M. Bhutta and M.~Liu, ``Pcr-pro: 3d sparse and different scale point
  clouds registration and robust estimation of information matrix for pose
  graph slam,''

\bibitem{davison2007monoslam}
A.~J. Davison, I.~D. Reid, N.~D. Molton, and O.~Stasse, ``Monoslam: Real-time
  single camera slam,'' {\em IEEE Transactions on Pattern Analysis \& Machine
  Intelligence}, no.~6, pp.~1052--1067, 2007.

\bibitem{klein2007parallel}
G.~Klein and D.~Murray, ``Parallel tracking and mapping for small ar
  workspaces,'' in {\em Proceedings of the 2007 6th IEEE and ACM International
  Symposium on Mixed and Augmented Reality}, pp.~1--10, IEEE Computer Society,
  2007.

\bibitem{newcombe2011dtam}
R.~A. Newcombe, S.~J. Lovegrove, and A.~J. Davison, ``Dtam: Dense tracking and
  mapping in real-time,'' in {\em 2011 international conference on computer
  vision}, pp.~2320--2327, IEEE, 2011.

\bibitem{engel2014lsd}
J.~Engel, T.~Sch{\"o}ps, and D.~Cremers, ``Lsd-slam: Large-scale direct
  monocular slam,'' in {\em European conference on computer vision},
  pp.~834--849, Springer, 2014.

\bibitem{forster2014svo}
C.~Forster, M.~Pizzoli, and D.~Scaramuzza, ``Svo: Fast semi-direct monocular
  visual odometry,'' in {\em 2014 IEEE international conference on robotics and
  automation (ICRA)}, pp.~15--22, IEEE, 2014.

\bibitem{mur2017orb}
R.~Mur-Artal and J.~D. Tard{\'o}s, ``Orb-slam2: An open-source slam system for
  monocular, stereo, and rgb-d cameras,'' {\em IEEE Transactions on Robotics},
  vol.~33, no.~5, pp.~1255--1262, 2017.

\bibitem{engel2017direct}
J.~Engel, V.~Koltun, and D.~Cremers, ``Direct sparse odometry,'' {\em IEEE
  transactions on pattern analysis and machine intelligence}, vol.~40, no.~3,
  pp.~611--625, 2017.

\bibitem{hata2014road}
A.~Hata and D.~Wolf, ``Road marking detection using lidar reflective intensity
  data and its application to vehicle localization,'' in {\em 17th
  International IEEE Conference on Intelligent Transportation Systems (ITSC)},
  pp.~584--589, IEEE, 2014.

\bibitem{wolcott2017robust}
R.~W. Wolcott and R.~M. Eustice, ``Robust lidar localization using
  multiresolution gaussian mixture maps for autonomous driving,'' {\em The
  International Journal of Robotics Research}, vol.~36, no.~3, pp.~292--319,
  2017.

\bibitem{wolcott2014visual}
R.~W. Wolcott and R.~M. Eustice, ``Visual localization within lidar maps for
  automated urban driving,'' in {\em 2014 IEEE/RSJ International Conference on
  Intelligent Robots and Systems}, pp.~176--183, IEEE, 2014.

\bibitem{lu2017monocular}
Y.~Lu, J.~Huang, Y.-T. Chen, and B.~Heisele, ``Monocular localization in urban
  environments using road markings,'' in {\em 2017 IEEE Intelligent Vehicles
  Symposium (IV)}, pp.~468--474, IEEE, 2017.

\bibitem{kim2018stereo}
Y.~Kim, J.~Jeong, and A.~Kim, ``Stereo camera localization in 3d lidar maps,''
  in {\em 2018 IEEE/RSJ International Conference on Intelligent Robots and
  Systems (IROS)}, pp.~1--9, IEEE, 2018.

\bibitem{Lujia2012Towards}
L.~Wang, M.~Liu, M.~Q.-H. Meng, and R.~Siegwart, ``Towards real-time
  multi-sensor information retrieval in cloud robotic system,'' in {\em
  Proceedings of the IEEE Conference on Multisensor Fusion and Integration for
  Intelligent Systems (MFI)}, 2012.

\bibitem{7403967}
L.~Wang, M.~Liu, and M.~Q.~H. Meng, ``A hierarchical auction-based mechanism
  for real-time resource allocation in cloud robotic systems,'' {\em IEEE
  Transactions on Cybernetics}, vol.~47, pp.~473--484, Feb 2017.

\bibitem{8772088}
B.~{Liu}, L.~{Wang}, and M.~{Liu}, ``Lifelong federated reinforcement learning:
  A learning architecture for navigation in cloud robotic systems,'' {\em IEEE
  Robotics and Automation Letters}, vol.~4, no.~4, pp.~4555--4562, 2019.

\bibitem{7060735}
L.~Wang, M.~Liu, and M.~Q.~H. Meng, ``Real-time multisensor data retrieval for
  cloud robotic systems,'' {\em IEEE Transactions on Automation Science and
  Engineering}, vol.~12, pp.~507--518, April 2015.

\bibitem{medioni2000tensor}
G.~Medioni, C.-K. Tang, and M.-S. Lee, ``Tensor voting: Theory and
  applications,'' in {\em Proceedings of RFIA}, vol.~2000, 2000.

\bibitem{liu2014efficient}
M.~Liu, ``Efficient segmentation and plane modeling of point-cloud for
  structured environment by normal clustering and tensor voting,'' in {\em 2014
  IEEE International Conference on Robotics and Biomimetics (ROBIO 2014)},
  pp.~1805--1810, IEEE, 2014.

\bibitem{liu12robio}
M.~Liu, F.~Pomerleau, F.~Colas, and R.~Siegwart, ``{Normal Estimation for
  Pointcloud using GPU based Sparse Tensor Voting},'' in {\em IEEE Int. Conf.
  on Robotics and Biomimetics (ROBIO)}, 2012.

\bibitem{von2008lsd}
R.~G. Von~Gioi, J.~Jakubowicz, J.-M. Morel, and G.~Randall, ``Lsd: A fast line
  segment detector with a false detection control,'' {\em IEEE transactions on
  pattern analysis and machine intelligence}, vol.~32, no.~4, pp.~722--732,
  2008.

\bibitem{ballard1981generalizing}
D.~H. Ballard, ``Generalizing the hough transform to detect arbitrary shapes,''
  {\em Pattern recognition}, vol.~13, no.~2, pp.~111--122, 1981.

\bibitem{zhang2019vr}
J.~Zhang, L.~Tai, P.~Yun, Y.~Xiong, M.~Liu, J.~Boedecker, and W.~Burgard,
  ``Vr-goggles for robots: Real-to-sim domain adaptation for visual control,''
  {\em IEEE Robotics and Automation Letters}, vol.~4, no.~2, pp.~1148--1155,
  2019.

\bibitem{liu12navigation}
M.~Liu, C.~Pradalier, F.~Pomerleau, and R.~Siegwart, ``{The role of homing in
  visual topological navigation},'' in {\em 2012 IEEE/RSJ International
  Conference on Intelligent Robots and Systems, 2012.(IROS 2012)}, 2012.

\end{thebibliography}
	\bibliographystyle{ieeetr}

\end{document}